\documentclass[letterpaper, 10 pt, conference]{ieeeconf}  

\IEEEoverridecommandlockouts                              

\usepackage{graphicx}
\usepackage{wrapfig}
\usepackage{booktabs}
\usepackage{subcaption}
\usepackage{url}
\usepackage{xspace}
\usepackage{soul}

\usepackage[colorinlistoftodos,prependcaption,textsize=tiny]{todonotes}

\newcommand{\early}{\textsc{Early Fusion}\xspace}
\newcommand{\concat}{\textsc{Late Fusion}\xspace}
\newcommand{\late}{\textsc{Attention Map}\xspace}
\newcommand{\softmap}{\textsc{Softmax Attention Map}\xspace}

\title{\early{} for Goal Directed Robotic Vision}
\author{Aaron Walsman$^1$\ \ \ \ \  Yonatan Bisk$^1$\ \ \ \ \   Saadia Gabriel$^1$\ \ \ \ \ \  Dipendra Misra$^2$\\ Yoav Artzi$^2$\ \ \ \ \ \ \  Yejin Choi$^1$\ \ \ \ \ \ \  Dieter Fox$^{1,3}$\\
\thanks{$^{1}$Paul G. Allen School of Computer Science and Engineering, University of Washington.\ \ \ \ \ \ \ \ $^{2}$Cornell University.\ \ \ \ \ \ \ \  $^{3}$NVIDIA.}
}

\begin{document}

\maketitle
\thispagestyle{empty}
\pagestyle{empty}

\begin{abstract}
Building perceptual systems for robotics which perform well under tight 
computational budgets requires novel architectures which rethink the traditional computer vision pipeline. 
Modern vision architectures require the agent to build a summary representation of the entire scene, even if most of the input is irrelevant to the agent's current goal.
In this work, we flip this paradigm, by introducing \textsc{EarlyFusion} vision models that condition on a goal to build custom representations for downstream tasks. 
We show that these goal specific representations can be learned more quickly, are substantially more parameter efficient, and more robust than existing attention mechanisms in our domain.
We demonstrate the effectiveness of these methods on a simulated 
item retrieval problem that is trained in a fully end-to-end manner via imitation learning. 
\end{abstract}

\section{Introduction}
Robotics has benefited greatly from advances in computer vision, but sometimes the objectives of these fields 
have been misaligned.
While the goal of a computer vision researcher is often ``tell me what you see," the roboticist's is ``do what I say."
In goal directed tasks, most of the scene is a distraction.
When grabbing an apple, an agent only needs to care about the 
table or chairs if 
they interfere with accomplishing the goal.  Additionally, when a robot learns through 
grounded interactions,  architectures must be sample efficient in order to learn visual representations 
quickly for new environments.  
In this work we show how inverting the traditional perception pipeline: Vision $\rightarrow$ Scene Representation + Goal $\rightarrow$ Action
to incorporate goal information early into the visual stream allows agents to jointly reason and perceive:
Vision + Goal $\rightarrow$ Action, yielding faster and more robust learning.

We focus on retrieving objects in a 3D environment as an example domain for testing our vision architectures.
This 
task includes vocabulary learning, navigation, and scene understanding.  Task completion requires computing action trajectories and resolving 3D occlusions from a 2D image which satisfy the
user's requests. 
Fast and efficient planners work well in the presence of ground-truth knowledge 
of the world \cite{srinivasa2016system}.  However, in practice, this ground-truth knowledge is difficult to obtain, and we must often settle for noisy estimates.
Additionally,
when many objects need to be collected or moved, the planning problem search space grows rapidly. 

Unlike computationally expensive modern vision algorithms, we are interested in training perception algorithms with a more natural source of supervision, example demonstrations and imitation learning, in lieu of expensive large scale collections of labeled images.  This is particularly important for developing agents that learn new object classes on the fly (e.g. when being integrated into a new environment).  
Our work is most closely related to recent advances in instruction following and visual attention \cite{Misra:2017aa, Misra:18goalprediction}, but we do not provide explicit supervision for object detections or classifications. Finally, we will make the assumption that goals are specified by a simple list of object IDs, so as to avoid the ambiguity introduced by natural language commands.\\

\noindent \textbf{Contributions:} We show that early fusion of goal information in the visual processing pipeline (\early{})
outperforms traditional approaches and learns faster.
Furthermore, model accuracy does not degrade in performance
even when reducing model parameters by orders of magnitude (from 6M to $\sim$25K).

\section{Task Definition}
\begin{figure}[t]
\centering
\includegraphics[trim={5pt 0 5pt 0}, width=\linewidth]{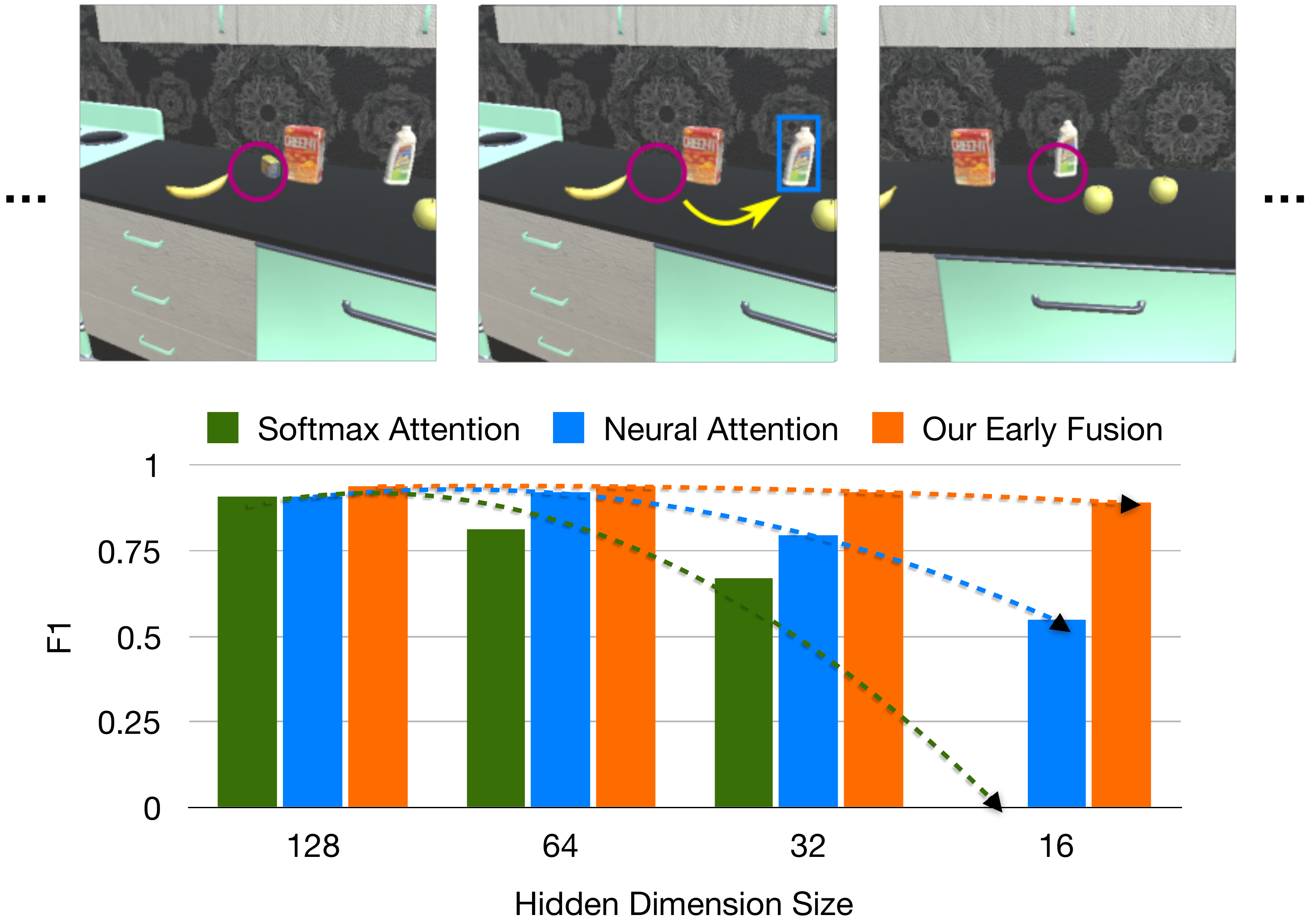}
\caption{We introduce a novel neural architecture for goal directed object detection which we demonstrate in a 
simulated table clearing task shown in the top row.  We
demonstrate that unlike conventional approaches, this
structure is stable under extreme parameter budgets
as seen in the bottom row.
}
\label{fig:episode_seq}
\end{figure}


In order to test the performance of \early
we built a simulated robotic task in which the objective
is to collect objects in a 3D scene as efficiently as possible.  The
agent is presented with a cluttered scene and a list of requested objects.  Often 
there are multiple instances of the same object, and there
can be unrequested objects blocking
the agent's ability to reach a target.  This forces the agent to reason about which object is closest
and remove obstructions as necessary.  The list of requested objects that remain in the scene
is presented to the agent at every time step, to avoid conflating scene understanding performance with issues of memory. 
The goal (Fig. \ref{fig:episode_seq} and
\ref{fig:occlusion}) is to train an agent to
optimally collect a list of objects from a
cluttered counter.

\subsection{Simulation Environment: CHALET}

\begin{wrapfigure}[13]{r}{0.5\linewidth}
\vspace{-10pt}
\centering
\includegraphics[width=\linewidth]{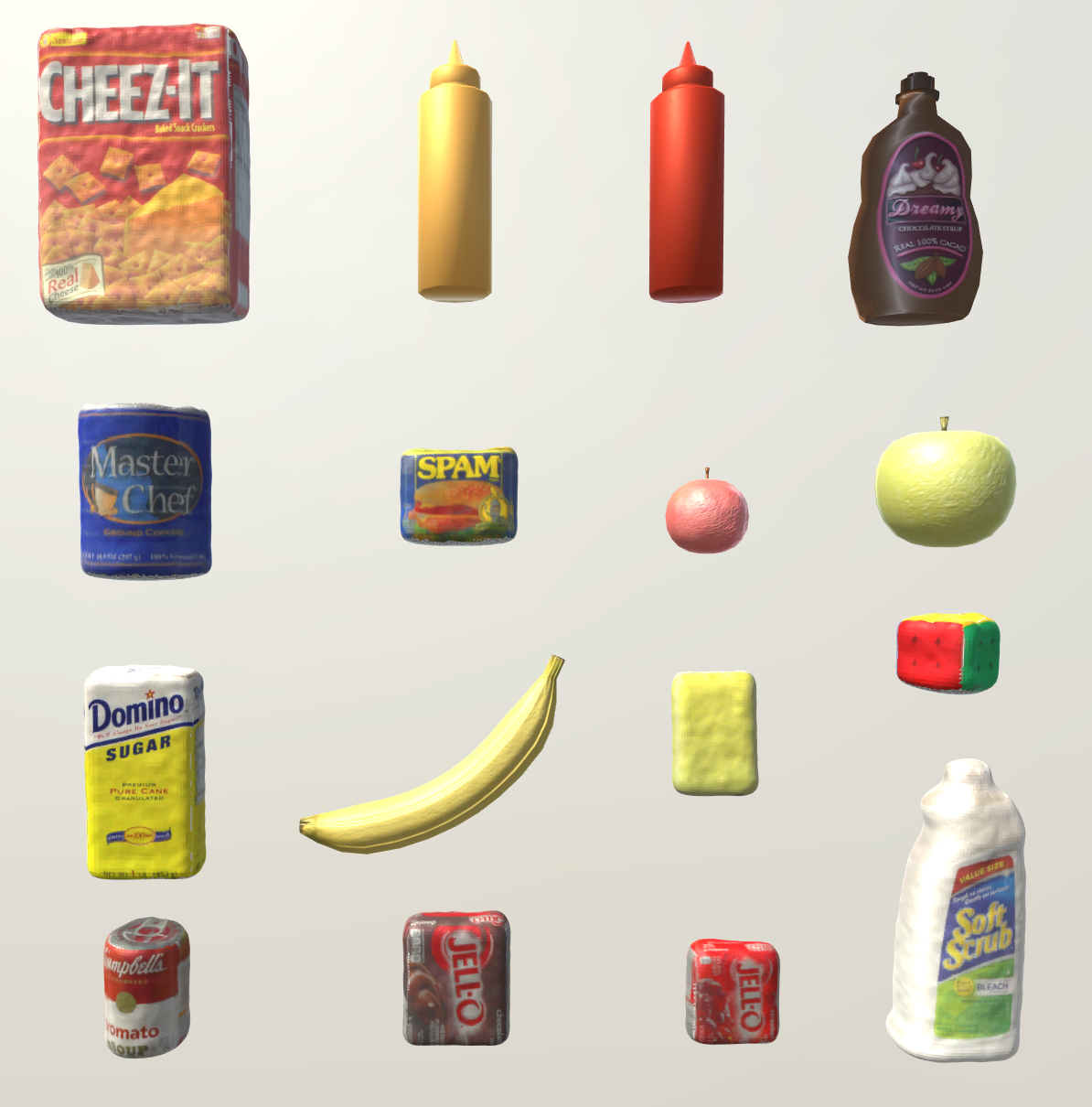}
\caption{Our 16 object types.}
\label{fig:object_types}
\end{wrapfigure}

Our environment consists of a tabletop setting with randomly placed objects, 
within a kitchen from the CHALET \cite{Yan:18} house environment.
Every episode consists of a randomly sampled environment 
which determines the set of objects 
(number, position, orientation and type) in addition to which
subset will be requested.  When there is more than one instance 
of a particular object, collecting any instance will satisfy the 
collection criteria, but one may be closer and require fewer steps to reach. 
Fig. \ref{fig:object_types} shows the sixteen object
types that we use for this task (six from CHALET and ten from the YCB dataset). Importantly, these are common household items, many of which cannot be detected by off the shelf ImageNet trained models.

The objects are chosen randomly and placed at a random location (x,y) on the table with 
a random upright orientation ($\theta$).  Positions and orientations are sampled until a 
non-colliding configuration is found. A random subset of the instances on the table are used for
the list of requested objects.  This process allows the same object type to be requested 
multiple times if multiple of those objects exist in the scene. Additionally, 
random sampling means an object may serve as a target in one episode and a distractor in the next. 
The agent receives 128x128 pixel images of the world and has a 60$^{\circ}$ horizontal field of view, requiring some exploration if a requested object is not in view.
\begin{wrapfigure}[17]{r}{0.5\linewidth}
\centering
\includegraphics[width=0.85\linewidth]{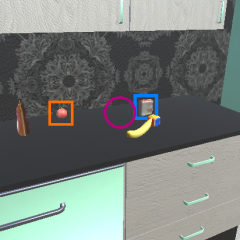}
\caption{Collecting the Jello (\textcolor{blue}{blue} box) requires more steps than the peach (\textcolor{orange}{orange} box) due to occluding objects.
An object may be collected if it is within the magenta circle.}
\label{fig:occlusion}
\end{wrapfigure}

Our agent consists of a first-person camera that can tilt up and down and pan left and right
with additional \texttt{collect}, \texttt{remove} and \texttt{idle} actions.  Each of the pan and tilt actions 
deterministically rotate the camera 2$^{\circ}$ in the specified direction. The \texttt{collect}
action removes the nearest object that is within 3$^{\circ}$ of the center axis of the camera
and registers the object as having been collected for the purposes of 
calculating the agent's score.  This region is visualized in Fig. \ref{fig:occlusion} as a magenta circle in the center
of the frame.
The \texttt{remove}
action does the same thing as \texttt{collect}, but does not register the item as having been collected.
This is used to remove superfluous items occluding the requested target. 
Finally, the \texttt{idle} action performs no action and should only 
be used once all requested items have been collected.
All actions require one time step, therefore objects
which are physically closer to the center of the camera may take more time steps to
reach if they are occluded.  For example, in Fig. \ref{fig:occlusion}
the peach (\textcolor{orange}{orange} box) requires fewer steps to collect than the
Jello box (\textcolor{blue}{blue} box) because the banana and Rubik's cube must be
removed first.
The precision required to successfully
collect an object makes this a difficult task to master from visual data alone.

\section{Models}
In our task, models must learn to ground visual representations of the world to the description of what to collect.  How to best combine this information is a crucial modelling decision.
Most multimodal approaches compute a visual feature map representing the contents 
of the entire image before selectively filtering
based on the goal.  This is commonly achieved using
soft attention mechanisms developed in the language \cite{Bahdanau:2014aa} and vision
\cite{Ba:2014aa, Mnih:2014aa, singh2018attention} communities.  

Attention re-weights the image representation and leads to more informative gradients, helping
models learn quickly and efficiently.
Despite its successes, attention has important limitations 
Most notably, because task specific knowledge is only incorporated
late in the visual processing pipeline, the model must first build dense image representations that
encode anything the attention might want to extract 
for all possible future goals. 
In complex scenes and tasks, this places 
a heavy burden on the initial stages of the vision system.
In contrast, we present a technique that injects goal information early into the visual 
pipeline in order to build a task specific representation of the image
from the bottom up.
Our approach avoids the traditional bottleneck imposed on perception systems, and allows the model to
discard irrelevant information immediately.
Our system may still need to reason about multiple objects in
the case of clutter and occlusion (e.g. target vs distractor), but its perception can ignore all
objects and details that are not relevant for the current task.

\begin{figure*}[t]
\centering
\includegraphics[width=0.95\linewidth]{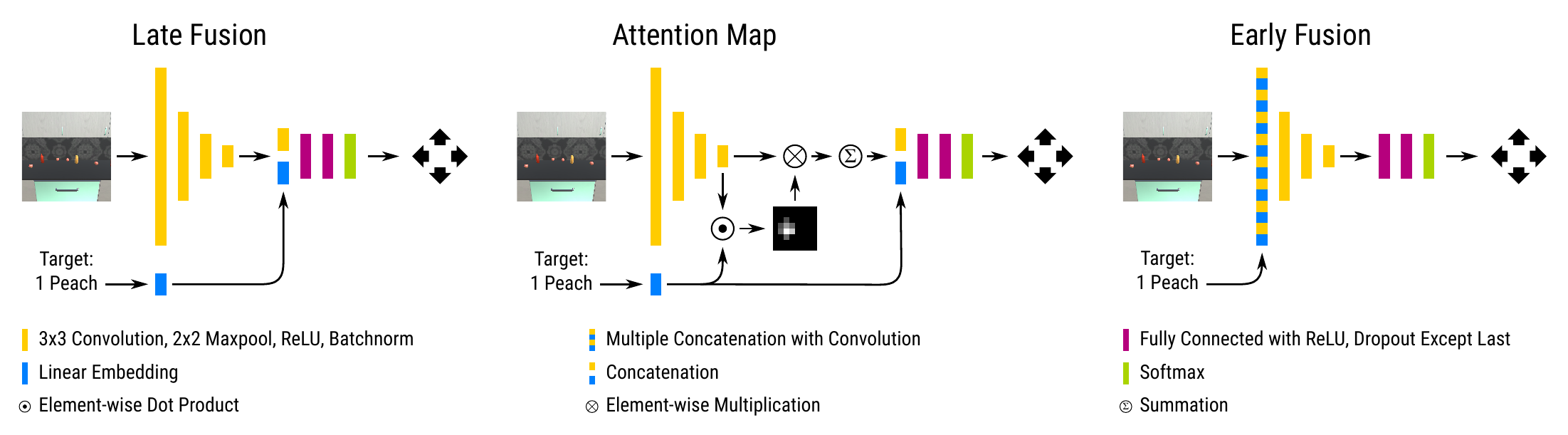}
\vspace{-5pt}
\caption{We compare a simple concatenation of visual and goal representations (\concat), against two variations of the attention mechanism above, and \early{} to isolate the effects of \textbf{when} multimodal representations are formed.}
\label{fig:networks}
\end{figure*}

Below, we briefly describe the three models (Figure \ref{fig:networks}) we compare:  Traditional approaches with delayed goal information (\concat{} \& \late{}) versus our goal conditioned \early{} architecture.

\subsection{\concat{}}
\concat{} constructs a single holistic representation of the entire
image via a stack of convolution and pooling layers before concatenating an embedding of
the requested objects in order to predict an action.
An object embedding is computed using a simple linear layer designed to turn a one-hot
encoding of the object into a dense representation. The complete request for multiple
objects is computed as a sum of these individual object embeddings.
This design forces the vision module to store 
semantic and spatial information about every object in the scene so
the final fully connected layers can ground target objects and reason about actions.

\subsection{\late{}}
We test traditional attention mechanisms over image regions.
As with \concat{}, the first step of this model is to pass the image through a
stack of convolution layers.
Rather than concatenate the request embedding directly onto the resulting
representation, these models first compute an attention map over the spatial
dimensions of the convolution output.
This is accomplished by comparing the embedded target vector with each
region of the convolutional feature map via a simple dot product.
This provides a weight to each region which can
then be used to form the final image representation 
$I = \sum_i \frac{\alpha_i}{Z} h_i$.  Next, $I$ is 
concatenated to the request to make an action decision.
We test two attention models: \softmap{} which is
defined above and \late{} which is unnormalized.  Using a
softmax leads to a peakier distribution which focuses the model on fewer regions of the image (see Fig. \ref{fig:attn_maps}).

In contrast to
the \concat{} model, the attention mechanism provides a filter on extraneous 
aspects of the image to simplify the control processing.  In these models the grounding
from image features to goal objects is done with a direct comparison operator
(the dot product).
These models are widely used for Visual Question Answering (VQA) problems on static images.  We also explored 
more complex models \cite{singh2018attention} for computing attention maps, 
but found this traditional version worked the best in our setting
and provided a strong baseline for comparison. 
In our results, we follow \cite{liu2018intriguing} and append spatial grids to the first layer of this network to encode spatial knowledge.   This extra information proved necessary for the attention models to compete with \early{}.\footnote{Spatial grids did not aid nor hinder \concat{}.}

\subsection{\early{}}

\begin{wrapfigure}[15]{r}{0.5\linewidth}
\centering
\vspace{-15pt}
\includegraphics[width=\linewidth]{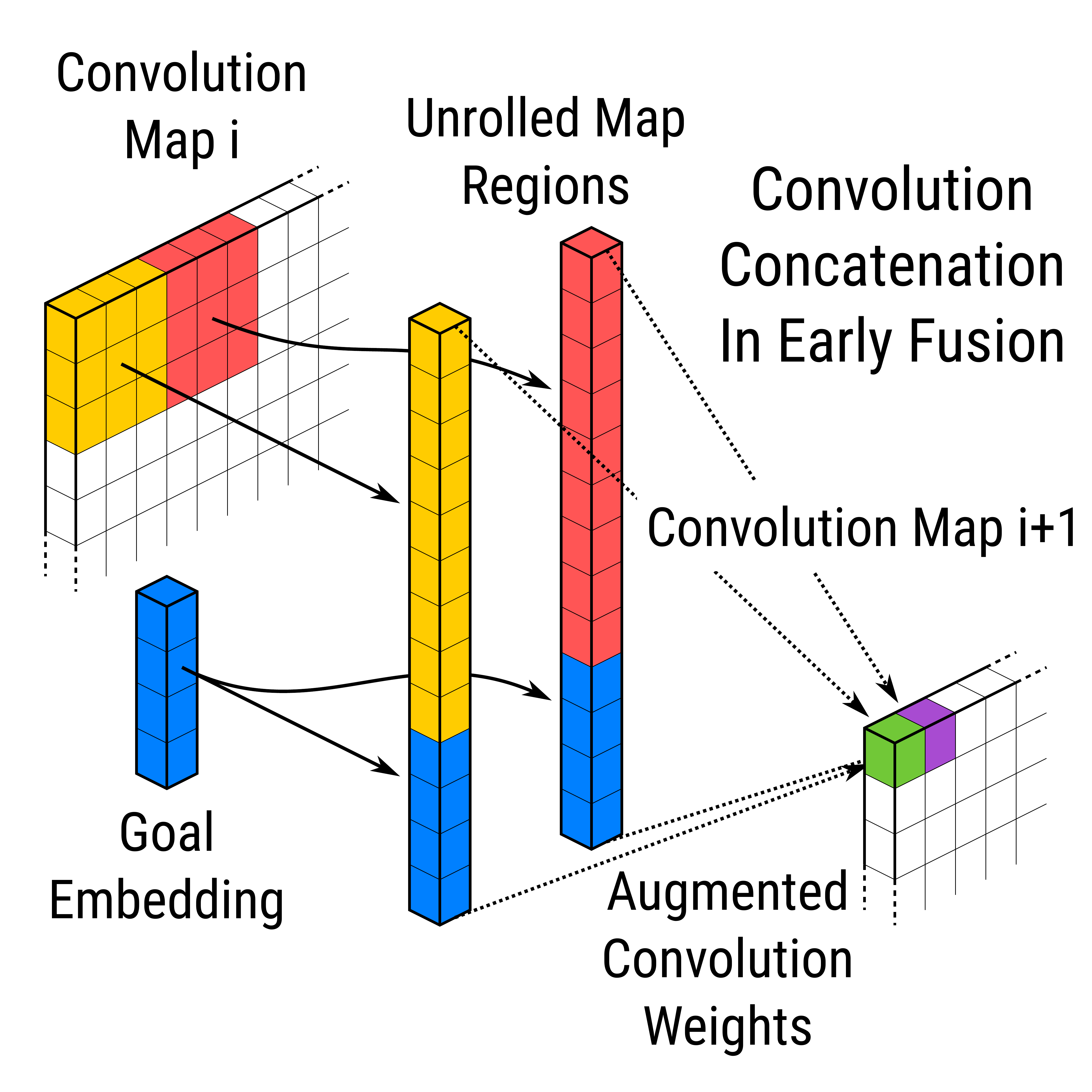}
\caption{\early{} of goal information with visual data.
The goal is concatenated with each block of visual data. 
}
\label{fig:conv_concat}
\end{wrapfigure}

Finally, our \early{} approach concatenates the request 
embedding to every region of a convolutional filter map.
This feature is then processed normally by a set of convolution kernels that
have been augmented to account for the extra channels.
Fig. \ref{fig:conv_concat} shows this process.
All further processing in the network is computed normally.
The model's subsequent convolution and fully connected layers may filter the visual
information according to the goal description that is now combined with the visual input. 
This results in an image representation which
contains only the necessary 
information for deciding the next action, effectively gaining the benefits of
a bottleneck while dispersing the logic throughout the network.
Critically, this means that the network does not have to build a semantic representation
of the entire image (See section \ref{sec:info_retention} for details).

Two important results of this architecture are:
1. Because the goal information is incorporated early, the network can learn to ground
the image features to the goal objects at any point in the model 
without additional machinery (like attention); and 2. 
The model can compute and retain the spatial information needed for its next action
without requiring the addition of a spatial grid.
These benefits allow us to obviate the complexity of other approaches, minimize parameters, and outperform other approaches on our task.

\subsection{Imitation Learning}
All models are trained with imitation learning using an oracle with direct access to the
simulator's state.  Similar to
DAgger~\cite{Ross:2011} and Scheduled
Sampling~\cite{bengio2015scheduled} we use an alternating
two-step training process.  In the first step, we roll out trajectories using the current model
while collecting supervision from the expert.  In the second step we use batches of recent trajectories
to train the model for a small number of epochs.
We then repeat this process and collect more data with the improved policy\cite{Blukis:18drone}.
We found that for our item retrieval problem this was faster to train than a more faithful
implementation of DAgger which would train a new policy on all previous data at each step,
and offered significant improvements over behavior cloning (training on trajectories demonstrated
by the expert policy).\footnote{Collecting 50 traj. in each roll-out
step, and training on the most recent 150 traj. for
three epochs in each training step produced the best results.}



Rather than teach our agents to find the shortest path to multiple objects, which is intractable in
general, we design our expert policy to behave greedily and move to collect the requested object
that would take the fewest steps to reach (including the time necessary to remove occluding objects).

\subsection{Implementation Details}
Since our goal is to construct a lightweight network that
is fast to train and evaluate, we use a simple image processing
stack of four convolution layers.  While this is small relative
to models trained for state-of-the-art performance on real images,
it is consistent with other approaches in simple simulated settings
\cite{santoro2017simple}.
All convolutions have 3$\times$3 kernels with a padding of one, followed
by 2$\times$2 max-pooling, a ReLU nonlinearity~\cite{glorot2011deep} and batch normalization~\cite{Ioffe:2015aa}.  This means each layer
produces a feature map with half the spatial dimensions of the
input.  The convolution layers are followed by two fully
connected layers, the first of which uses a ReLU nonlinearity
and Dropout \cite{Srivastava:2014aa} and the second of which
uses a softmax to produce output controls.
The number of convolution channels and hidden dimensions in the
fully connected layers vary by experiment
(see Section~\ref{sect:network_capacity}).
\footnote{
All of our models are optimized with Adam~\cite{Kingma:2015} with a learning rate of 1$e$-4, 
and trained with dropout~\cite{Srivastava:2014aa}.
The training loss was
computed with cross-entropy over the action space. All models and code were written in
PyTorch~\cite{paszke2017automatic} and will be made available.
}

\paragraph{Images} 
Our images are RGB and 128x128 pixels, but as is common practice in visual
episodic settings~\cite{mnih2013playing} we found our models performed best 
when we concatenated the most recent three frames to create a 9x128x128 input.
\footnote{Frames are black when they are not available
in the first two frames.}

\paragraph{Requests}
Models are provided the remaining items to collect as a list of one-hot
vectors.
Each of these items is passed through a learned embedding (linear) layer to produce an encoding.
These are then summed to produce a single
dense vector ($Target$).
Because the sequence order is not important to our
task, we found no benefit from RNN based encodings, though the use of an embedding layer, 
rather than a count vector, proved essential to model performance.

\begin{figure}[t]
\centering
\includegraphics[trim={11pt 0 11pt 0},width=0.96\linewidth]{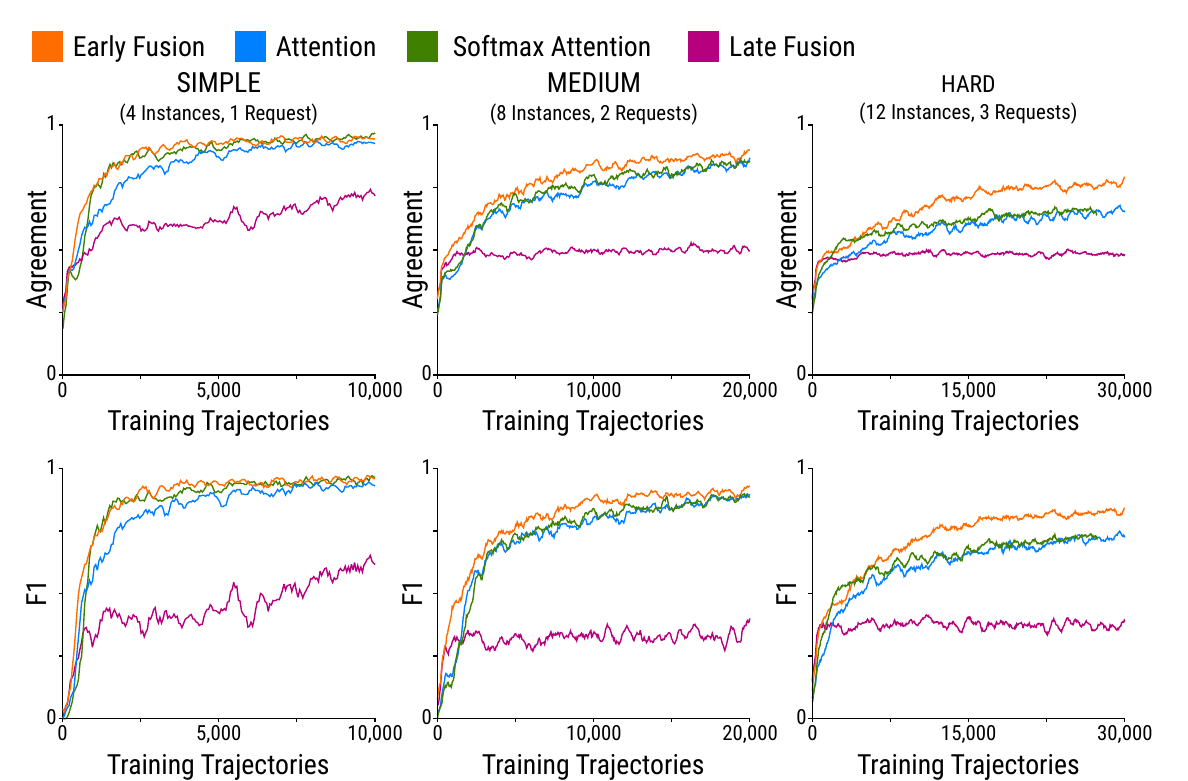}
\caption{Model performance on \textsc{Simple}, \textsc{Medium}, and \textsc{Hard} learning paradigms.  Models were run to convergence. 
}
\label{fig:success_chart}
\vspace{-5pt}
\end{figure}

\begin{figure*}[t]
\begin{center}
\includegraphics[trim={5pt 0pt 7pt 0},clip,width=\linewidth]{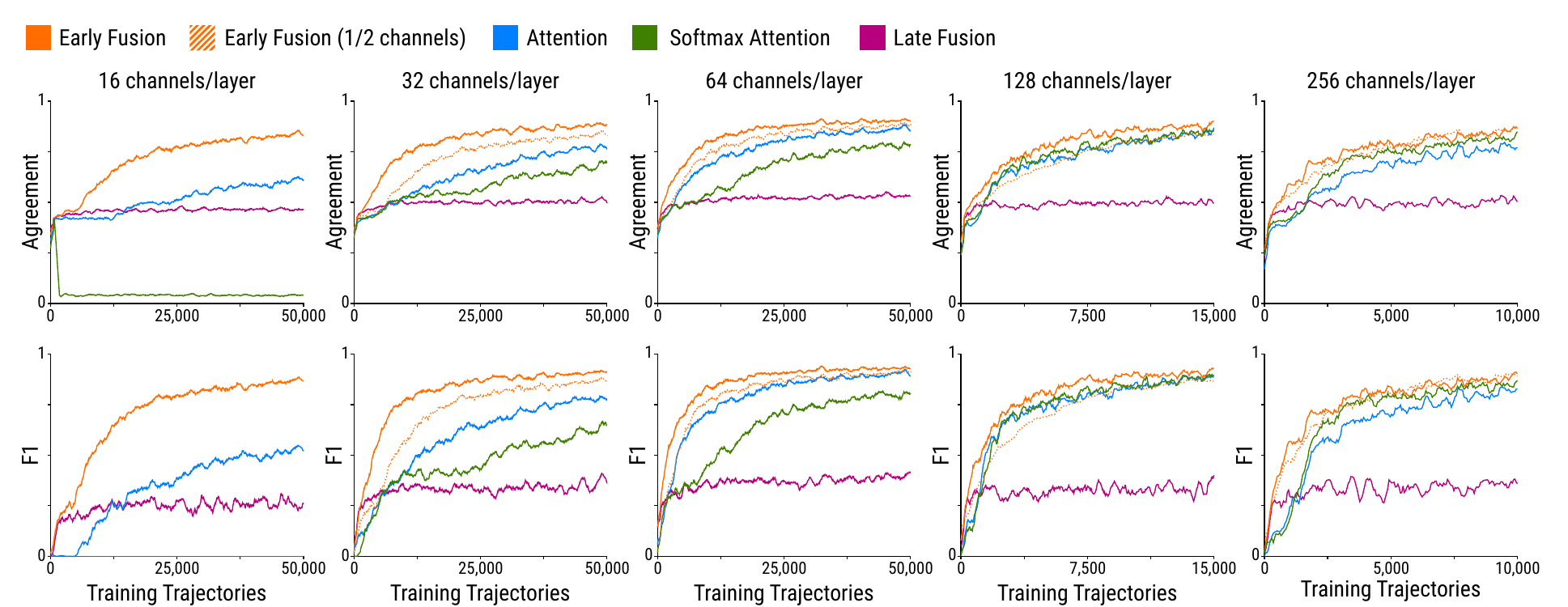}
\end{center}
\vspace{-5pt}
\caption{Ablation analysis showing the effect of the number of convolution channels and fully-connected hidden units on network performance.
Note that the scale of the x-axes in these plots varies due to longer training times for smaller networks to converge. Dashed \early lines plot the performance of the model with half the reported number of filters to include a comparison ensure where model has fewer parameters than the attention based approaches.
}
\label{fig:ablation}
\end{figure*}

\section{Experiments}


We tested all four models on a series
of increasingly cluttered and
difficult problems.  We also tested these
models with varying network capacity by
reducing the number of convolution channels
and features in the fully connected layers.
In all of these experiments, our \early
model performs as well or better than
the others, while typically training 
faster and with fewer parameters.

\subsection{Varying Problem Difficulty}

To test models on problems
of increasing difficulty, we built three
variations of the basic task by varying
clutter and the number of requested
items.
In the simplest task (\textsc{Simple}), each episode starts with four instances randomly placed on the
table and one object type is requested.  Next, for \textsc{Medium} eight instances are placed and two are
requested.  Finally, for \textsc{Hard} twelve instances are placed and three are requested.  Note that as the clutter
increases, the agent is presented with
not only a more complicated visual environment, but must also
work in a more complex action domain where it is increasingly
important to use the \texttt{remove} action to deal with
occluding objects.  The agent's goal
is to collect only the requested items in the allotted time.  
To evaluate peak performance for these experiments
we fixed the number of convolutions and
hidden layer dimensions in the fully
connected layers to 128.

Each episode runs for forty-eight steps, during which it is possible for the agent
to both successfully collect requested objects and erroneously collect items that were not requested.
We therefore measure task completion using
an F1 score.  Precision is the percentage of collected objects that were actually requested,
and recall is the percentage of requested objects that were collected.  The F1 score is computed at the
end of each episode.
In addition, we report overall agreement between the model and the expert's actions over the 
entire episode.  Figure \ref{fig:success_chart} plots the results of all four models on each of these problems
as a function of training time.



\begin{figure}[t]
\centering
\vspace{-10pt}
\includegraphics[trim={10pt 0 20pt 0}, width=\linewidth]{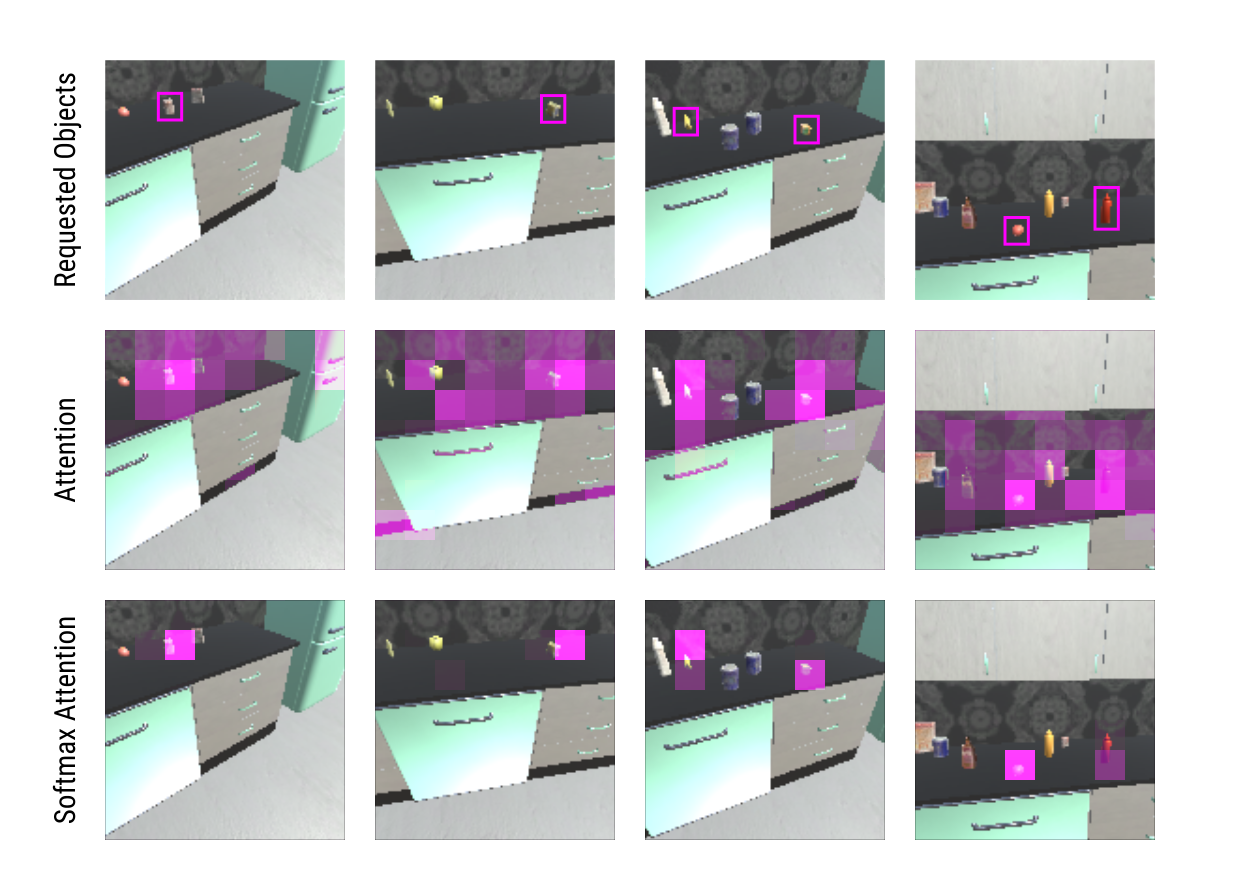}
\vspace{-20pt}
\caption{Attention visualizations for \late{} and \softmap{} models.  Targets are indicated here with magenta boxes in the top row.}
\label{fig:attn_maps}
\end{figure}

\paragraph{\textsc{Simple}} Except for the \concat{} model, which performs poorly in all scenarios, all models are able to
master the easiest task.  The \early{} and \softmap{} models learn quickly, but \late{}
eventually catches up to them.  The failure of the \concat{} baseline on this task shows
that even the simplest version of this problem is non-trivial.

\paragraph{\textsc{Medium}} The intermediate problem formulation is clearly more difficult, as no models are able to perform as well on
it as the easiest problem.  The \early{} model gains a small but significant improvement in performance
while \softmap{} and \late{} are slightly worse, but comparable to each other.

\paragraph{\textsc{Hard}} 
This case contains more cluttered images and more complex goal descriptions.  The \early{} model is clearly superior; it learns significantly faster
than the other models and results in higher
peak performance.

It is also worth comparing the \late{} and \softmap{} models.  While these models perform
similarly on these tasks, the \softmap{} model learns faster than the \late{} model on the easiest task,
but slightly slower on the more difficult ones.  We posit that the softmax focuses the attention heavily
on only a few regions, which is useful for sparse uncluttered environments, but less appropriate when the network
must reason about multiple objects in different regions.

Fig. \ref{fig:attn_maps} provides a comparison of attention maps. 
Unsurprisingly, the \softmap{} model produces a sharper distribution around the requested objects, but both methods
correctly highlight the objects of interest.  In this work, we have limited 
our definition of clutter to 12 items per scene, in part for ease of visualization
and compute time.  

\subsection{Varying Network Capacity}
\label{sect:network_capacity}
Having demonstrated that \early is
at least as powerful as 
attention based approaches while 
being simpler (no grid information or attention logic), we explore how these 
approaches perform on varying parameter budgets.  Real-time and embedded systems require 
efficiency both when training and during inference.
Since \early removes irrelevant information early in the processing pipeline,
we expect it to require less network capacity than the other methods.
To test this claim, we re-run our 
\textsc{Medium} difficulty setting (because attention models performed well) and 
compare performance when models have access to 256, 128, 64, 32, or only 16 channel
convolutions and fully connected layers, reducing our model sizes by several orders of magnitude.

In Fig. \ref{fig:ablation}, training time increases for small networks, 
but \early is able to quickly achieve around the same final
performance regardless of the extremely small network capacity.   This allows for dramatically more efficient inference
and parameter/memory usage.  In contrast, other models degrade substantially
as the network size decreases.
Note that after 50,000 trajectories it appears that attention based models are still 
slowly improving, but there is a stark contrast in learning rates.  In particular, 
for the smallest models (16) we see that \late, even after training for twice 
as long as \early, still has half the performance.

Because attention mechanisms collapse their final representations, they have a smaller fully connected layer 
and therefore fewer parameters for the same number of channels.  To account for this, we have also included 
a dashed orange line in Fig. \ref{fig:ablation} labeled early fusion (1/2 channels) which shows the performance of \early with half the channels as the other models and fewer parameters.  Again smaller \early networks
outperform and learn faster than the other approaches.



\subsection{Generalization}
To measure generalization 
we conduct experiments in which the agent is trained on a subset of the possible request combinations
and then tested on unseen requests.  Here the agent is trained with 128 different
two-item combinations, 
and then tested on a held out 128 two-item combinations (Rows 1 and 2 below).  In this setting, the agent generalizes to unseen item pairs, indicating that the agent is not merely memorizing these combinations, but learning to recognize the structure of requests composed of individual objects.

\begin{center}
\begin{tabular}{lcc}
        & Agreement & F1\\
        \midrule
        Two-Item Train & 0.8918 & 0.9215\\
        Two-Item Test & 0.8695 & 0.8938\\
        Three-Item Test & 0.8140 & 0.8243
\end{tabular}
\end{center}

In the second experiment, the same agent was tested on a random collection of three-item combinations to determine if
the agent can generalize to higher counts than during training (Row 3). 
The agent is surprisingly robust to this variation.

\subsection{Information Retention}
\label{sec:info_retention}
\begin{figure}[t]
\centering
\includegraphics[width=0.85\linewidth]{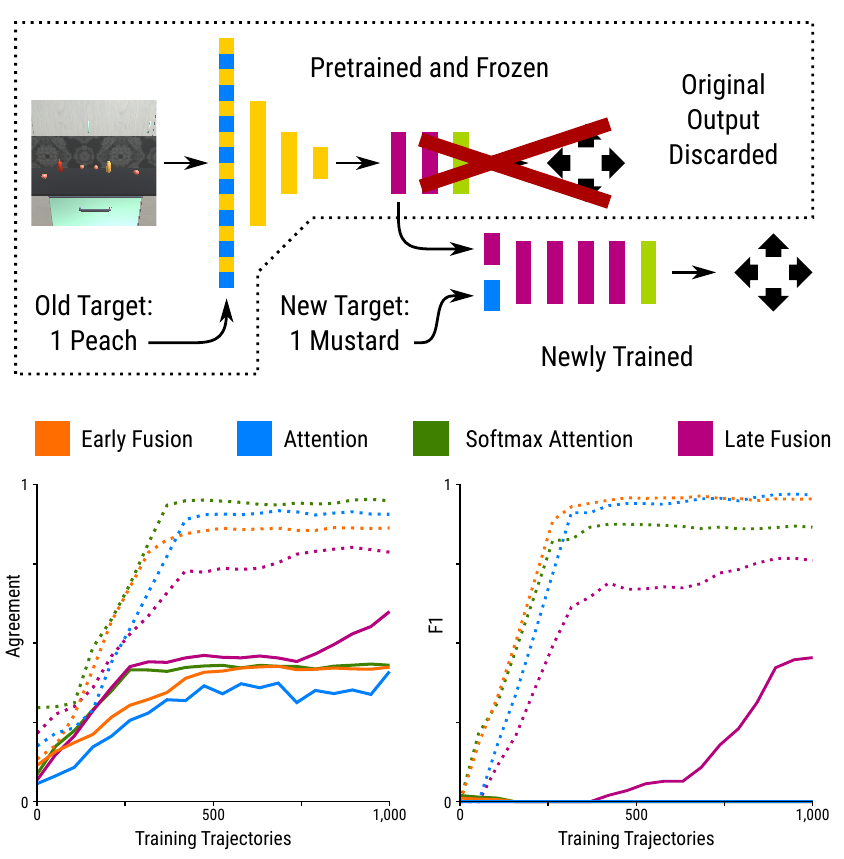}
\caption{The top is the frozen \early{} network with an new trainable branch 
for collecting an alternate test object.  On bottom is the performance of this approach when the old
target and the new target are aligned (dotted lines) and performance for when they are
different (solid lines).\protect\footnotemark
}
\label{fig:network_mismatch}
\end{figure}

We have argued above that knowing the request allows the network to
discard information about irrelevant objects in the scene.
To investigate how much information is retained in the intermediate stages of
the network we use the hidden states from models
trained on the \textsc{Simple} task and assess whether they can be used
to predict the correct action for a new query that is different than the one they were 
conditioned on.  This is implemented by freezing the original model,
and training a new set of final layers with a second conditional (Figure \ref{fig:network_mismatch}).
In this experiment, we use the \concat model as a proxy for the layer prior to attention in those models.

We find that if the same request is fed to both the original network
and the new branch, we achieve performance comparable
to the original model (dotted lines).  On the other hand if mismatched requests are fed into the two branches
all models suffer a substantial degradation of performance, with most unable to collect a
single object (solid lines).  Both Early Fusion and the attention models have completely removed irrelevant information,
while Late Fusion approaches appear to only retain much of the irrelevant information.

\footnotetext{Note $40\%$ agreement is the majority class baseline as movement is more common than collection.}

\section{Related Work}
Processing strategies for goal-directed visual search have been an important area
of study in psychology, neuroscience and computer vision for many years \cite{wolfe1994, TSOTSOS1995507}.
Early work in this area drew on the observation that human and primate vision seems to be
at least partially driven by goal-directed top-down signals
\cite{Frintrop:2010:CVA:1658349.1658355, Frintrop:2005:GST:2104486.2104505}.

More recently there has been a proliferation of works examining goal directed visual
learning in simulated worlds
\cite{Gordon:17,embodiedqa,Hermann:17,Anderson:17, codevilla2017end, Zhu:2017a} which each aim to 
bring different amounts of language, vision and interaction to the task of navigating a 3D environment.
This has also been attempted in real
3D environments \cite{gupta2017cognitive}.
Importantly, in contrast to our work, these approaches often pretrain as much of their networks as possible.
\cite{Hermann:17} do not pretrain for their RL based language learning.  Their work 
does not address learning with occulusion or larger vocabularies.  
In parallel, the robotics
literature has investigated grounding instructions directly to robotic control
\cite{Matuszek:2010, Tellex:2011, misra:rss14, Zhu:2017a,Blukis:18drone,Blukis:18visit-predict}, a domain where data is expensive to collect.

Training end-to-end visual and control networks \cite{Levin:2017}, has proven difficult due to
long roll outs and large action spaces.
Within reinforcement learning, several approaches for mapping natural language instructions to actions rely 
on reward shapping \cite{Misra:2017aa,Misra:18goalprediction} and imitation 
learning \cite{Blukis:18drone,Blukis:18visit-predict}.  Imitation learning has also proven effective for fine grained activities
like grasping \cite{Eppner:2009}, leading to state-of-the-art results on a broad set of tasks 
\cite{ijcai2017-676}.
The difficulty encountered in these scenarios emphasizes the need to explore new
methods for efficient learning of multimodal representations.  \cite{singh2018attention} explored attention model architectures, but do not include early fusion techniques.
Early fusion of goal information has shown promise with small observation spaces \cite{tai2017virtual},
but our work begins to explore this method for high-dimensional visual domains.
In this paper, we hope to provide the community with a missing analysis and insights into this approach and its power in interactive settings.


\section{Conclusion}
Goal directed computer vision is an important area for robotics research and efficient training of high performing models with minimal footprints are essential for in situ learning.  We take one step in this direction by showing how \early{} is ideal for the simulated robotic object retrieval task, and preferable to traditional attention based approaches.  Future work should investigate how our approach and analysis can be generalized to on-device learning.

\IEEEtriggeratref{19}  
\bibliographystyle{IEEEtran}
\bibliography{references}

\end{document}